\pdfoutput=1
\documentclass{article}

\usepackage[preprint]{neurips_2019}

\usepackage[utf8]{inputenc} % allow utf-8 input
\usepackage[T1]{fontenc}    % use 8-bit T1 fonts
\usepackage{hyperref}       % hyperlinks
\usepackage{url}            % simple URL typesetting
\usepackage{booktabs}       % professional-quality tables
\usepackage{amsfonts}       % blackboard math symbols
\usepackage{nicefrac}       % compact symbols for 1/2, etc.
\usepackage{microtype}      % microtypography
\usepackage{textcomp}
\usepackage{mathptmx}
\usepackage{xcolor}
\usepackage{bm}
\usepackage{subfig}
\usepackage{fancyhdr,graphicx,amsmath,amssymb}
\usepackage[ruled]{algorithm2e}

\graphicspath{{figs/}{./}}

\title{Function Preserving Projection for Scalable Exploration of High-Dimensional Data}

\author{
  Shusen Liu, Rushil Anirudh, Jayaraman J. Thiagarajan and Peer-Timo Bremer \\
  Lawrence Livermore National Laboratory\\
  7000 East Ave, Livermore, CA 94550\\
  \texttt{ \{liu42, anirudh1, jayaraman1, bremer5\}@llnl.gov} \\
}

\begin{document}
\maketitle

\begin{abstract}
%In this work, we propose function preserving projection (FPP), a new class of scalable linear projection techniques for finding interpretable structure in high-dimensional datasets. Instead of focusing on preserving the structure of high-dimensional points as traditional dimensionality reduction methods, we aim to find a 2D view of the data, in which human interpretable patterns of meaningful properties (that defined on the high-dimensional points, i.e., functions in a high-dimensional domain) are revealed. Such a visual-centric approach introduces new avenues for exploratory data analysis and allows efficient implementation that provides a scalable tool for exploring potentially unknown or unexpected relationships in complex and large high-dimensional data (e.g., beyond millions of samples and tens of thousands of dimensions).
We present \textit{function preserving projections} (FPP), a scalable linear projection technique for discovering interpretable relationships in high-dimensional data. 
Conventional dimension reduction methods aim to maximally preserve the global and/or local geometric structure of a dataset.
However, in practice one is often more interested in determining how one or multiple user-selected \textit{response} function(s) can be explained by the data. 
To intuitively connect the responses to the data, FPP constructs 2D linear embeddings optimized to reveal interpretable yet potentially non-linear patterns of the response functions.
More specifically, FPP is designed to (i) produce human-interpretable embeddings; (ii) capture non-linear relationships; (iii) allow the simultaneous use of multiple response functions; and (iv) scale to millions of samples. 
Using FPP on real-world datasets, one can obtain fundamentally new insights about high-dimensional relationships in large-scale data that could not be achieved using existing dimension reduction methods.

\end{abstract}

\section{Introduction}
% Discovery relationship between dependent and independent variable are essential for scientific discovery
% Model fitting can be less interpretable and can not utilize the human vision as a guide for pattern discovery
% p-value and supervised dimensionality reduction
% Comparison with linear regression

The rapid advances in both experimental and computational capabilities have resulted in a deluge of data being collected in all branches of science and engineering. 
Whether this data describes outputs of computer simulations or observations from experiments, it is imperative to uncover patterns and relationships in the resulting high-dimensional data to gain insights into the underlying phenomena.
Existing solutions for high dimensional analysis struggle to balance the complexity of an approach with the need to interpret the results.
For example, one may fit a linear model which is easily understood yet cannot recover non-linear patterns or deploy more general techniques, such as deep neural networks, which are more flexible but difficult to interpret. 
Similarly, simple visual encodings such as scatterplot matrices and parallel coordinates~\cite{inselberg1990parallel, carr1987scatterplot} are easy to read but do not scale to large dimensions.
Non-linear dimension reduction can sometimes address the dimensionality challenge at the cost of losing interpretability as axes lose their meaning.
Furthermore, many existing visualization techniques for high-dimensional data do not scale gracefully to large datasets.

% either rely on building unintuitive computational models (e.g., deep neural network) for automated analysis or use visual encodings, such as scatterplot matrices or parallel coordinate plots that are often not scalable 

% This naturally requires a flexible pattern discovery framework that allows human-in-the-loop exploration. 
% However, most existing solutions either rely on building unintuitive computational models (e.g., deep neural network) for automated analysis that not suitable for human exploration and evaluation, or involving visual encodings~\cite{inselberg1990parallel, carr1987scatterplot} and interactive visualization systems~\cite{buja1996interactive, liu2015visual} that is often not scalable to handle the overwhelming size and complexity of scientific data. 
% %majority of existing solutions rely heavily on machine learning models to facilitate automated analysis, thus rendering them 
%black-box and uninterpretable. 
%While such fundamental limitation has be somewhat addressed by the introduction of various visual encoding and visualization system that allow user interactive exploring high-dimensional data, the overwhelming size and complexity of scientific data can make most comprehension task infeasible.  

Instead, we need an approach that can deal with large data and non-linear relationships while remaining interpretable. 
Function preserving projections (FPP) achieve both objectives by focusing on specific aspects of data through the lens of user-selected \textit{response} functions. 
Response function(s) typically correspond to one or multiple diagnostic measurements or simulation outputs, defined at each data point, and can provide a powerful window into underlying patterns in the data.
More specifically, rather than aiming to preserve the entire neighborhood structure of high-dimensional data in hopes of finding interesting patterns, we deliberately search for the best linear projection, such that the chosen response functions create an interpretable pattern in the projected space.
An important consequence of focusing on the projected response function is that FPP inherently ignores domain variables and structures not pertinent to the response and does not require the domain to have a low intrinsic dimension. The latter is a key property in many applications, most notably simulation ensembles, where the domain is defined as a uniform sampling of a high dimensional hypercube. In these cases, dimension reduction is futile as -- by definition -- there exists no low dimensional structure in the domain one could discover and thus many of the traditional techniques do not apply.
We consider a pattern interpretable if it can be approximated by a chosen regressor and by adjusting the type and order of the regressor, i.e., polynomial vs.\ exponential, linear vs.\ non-linear, etc., we allow users to choose the complexity of the pattern they deem acceptable.
The key insight driving the development of FPP is the fact that interpreting non-linear embeddings is challenging yet humans are highly skilled in understanding non-linear patterns. 
By restricting the initial project to a linear map FPP preserves the interpretability of the resulting plots, i.e., axis labels, while non-linear regressors enable us to discover noisy and non-linear relationships.

Conceptually, FPP is a dual approach to kernel machines in machine learning, which employ non-linear, and often infinite-dimensional mappings to enable the use of simple linear models for complex data. 
From a visual exploration standpoint, we argue that the use of linear models in $2$D is not necessary since humans can still interpret more complex relationships. 
On the other hand, non-linear mappings of the data coordinates are not explainable, thereby making the subsequent analysis also highly opaque. 
Another crucial feature of FPP is that it supports much larger data than related techniques and enables unified analysis with multiple response functions, wherein a single $2$D projection is identified that jointly preserves all responses.
Finally, an often-overlooked challenge for view-finding or any pattern detection algorithm is that, in higher dimensions, it is challenging to qualitatively distinguish between meaningful structure and artifacts. 
This behavior can be directly attributed to the curse of dimensionality, where the sample sizes used are not sufficient to make meaningful inferences about the data. 
In the case of view-finding approaches like FPP, this manifests as overfitting, where one can almost always create a seemingly meaningful pattern given low enough sample counts and sufficiently high dimensions. 
To address this problem we argument the $2$D embeddings with the equivalent of a $p$-statistic that quantifies the likelihood of the given hypothesis (the observed pattern) to occur in a random function. 
This, for the first time, provides a quantitative and easy to interpret indicator on how reliable a given visualization is likely to be, which is crucial to confidently infer new insights. 
Using several case studies we show that FPP provides comprehensive insights that cannot be easily obtained using conventional techniques.

\section{Related Work}

Even though FPP produces linear projections, it is fundamentally different from existing linear dimension reduction techniques like principal component analysis (PCA)~\cite{jolliffe2011principal}.
Dimension reduction, as currently understood, is typically aimed at preserving global (PCA) or local neighborhood structures (locality preserving projection~\cite{he2004locality}) of the high-dimensional point geometry.
The resulting projection is then colored according to a response function in hopes of highlighting interesting relationships. 
However, when analyzing particular response functions the complete geometry of the point set may not be relevant and can even be detrimental by introducing ``spurious'' variations unrelated to the response of interest. 
Instead, FPP directly targets the response functions of interest and preserves only the aspects of the high dimensional geometry relevant to the problem.

% Despite the proposed method always produces a linear projection, 
% due to the optimization goal, it is fundamentally different from popular linear dimensionality reduction methods, e.g., principal component analysis~\cite{jolliffe2011principal},
% which emphasis on the finding the subspace that best preserves either global~\cite{jolliffe2011principal} or local neighborhood structures~\cite{he2004locality} of the high-dimensional point geometry.

% Instead, our goal is to visualize the structure of the high-dimensional points that drive the variation of the function defined on them. 
In this context, FPP is similar to cross decomposition approaches, such as canonical component analysis (CCA)~\cite{hardoon2004canonical} and partial least square (PLS)~\cite{chin1998partial}.
CCA aims to find the subspace that best aligns the domain and the range of a high-dimensional function.
However, this produces a subspace that is at most equal to the minimal dimension of either the domain and the range. 
Consequently, for scalar functions, one can only find a 1D subspace, rather than a 2D projection, and for more than two response functions a secondary projection must be added for visualization which results in an unoptimized projection. 
Partial least square~\cite{chin1998partial} does not have this limitation but is restricted to a linear regressor making it difficult to identify even simple nonlinear patterns, i.e.\ a circle (see Section~\ref{sec:result}).
Similarly, inverse sliced regression~\cite{li1991sliced}, which utilize an inverse regression formulation to reduce the dimension of the input with respect to the response (function), is also limited to linear correlation structure.

For visualization of the nonlinear structure of the high-dimensional point, many nonlinear dimensionality reduction techniques~\cite{maaten2008visualizing, Kruskal1964} have been proposed. 
However, while they can capture some intrinsic structure of a point sample it is difficult to connect the observed patterns back to the domain as the axes no longer have well-defined meaning and distances can be heavily distorted.
Furthermore, similar to the linear projections these techniques do not consider the response function in creating the projection.
As a result, a projection that explains the response well is due to a fortunate coincident rather than a deliberate design. 

The technique most similar to FPP is projection pursuit regression (PPR)~\cite{friedman1981projection}, which has been designed as a universal high-dimensional function approximator for regression tasks.
The PPR fits a linear combination of multiple $1D$ non-linear transformations of linear combinations of variables in the data. 
The non-linear mapping allows PPR to capture certain nonlinear patterns. 
Similar to FPP, the PPR formulation can also be considered as a dual of the kernel regression as explored in~\cite{donoho1989projection}.
However, designed as a function approximator, the performance of PPR improves as we increase the number of $1D$ nonlinear transformation components, which can lead to challenges in its interpretation.
With FPP, we instead directly fit a non-linear model in the 2D projected space, which not only allows intuitive visualization but also simplified the optimization process that allows us to efficiently scale the proposed technique to extremely large sample size and dimensions.

%However, FPP is the first approach to utilize these concepts for visualization purposes.
%\ptb{The last sentence sounds weak.} 

%%% Local Variables:
%%% mode: latex
%%% TeX-master: "fpp_arxiv"
%%% End:

\begin{figure}[htbp]
\centering
  \includegraphics[width=0.8\linewidth]{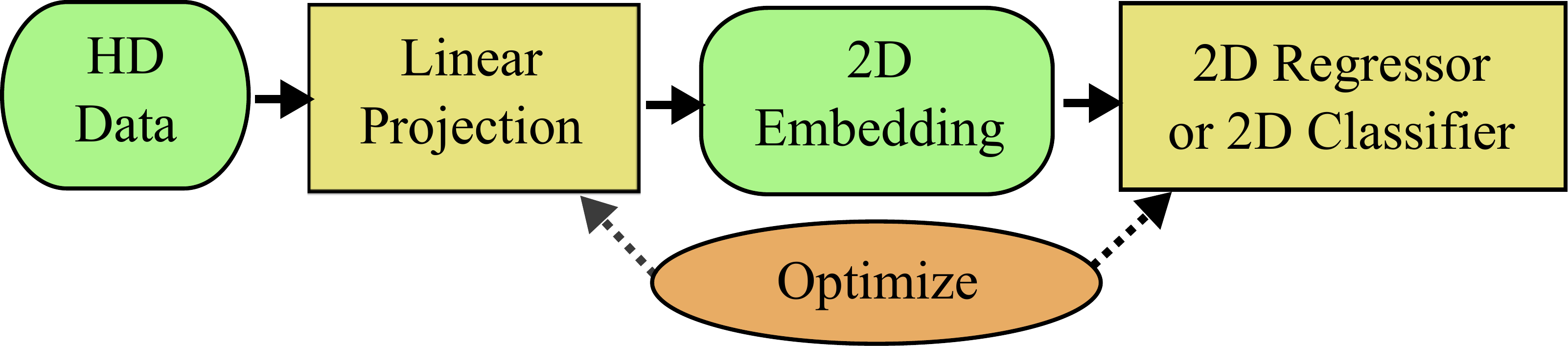}
 \caption{Overview of the computation pipeline for the proposed function preserving projection approach. We jointly optimize the linear projection and the 2D regressor (or classifier) for identifying human-interpretable pattern in high-dimensional functions.}
\label{fig:overview}
\end{figure}

\section{Method}
\label{sec:method}
As discussed in the introduction, the growing need for exploring large and complex high-dimensional dataset call for visualization tools that 1) produce interpretable embedding; 2) capable for capturing nonlinear pattern; 3) scalable to large sample size and high dimensionality.
For interpretability, contrary to many high-dimensional data visualization approach (e.g., t-SNE or MDS) that employ a complex map from high-dimensional to 2D space, we focus on a simple linear transformation that produces a 2D embedding with well-defined axes.
To capture the potentially nonlinear structure of the function, we frame the pattern discovery as a nonlinear 2D regression problem, where the choice of the regressor, i.e., the polynomial degree, provides direct control over the visual complexity a user expects or is willing to consider as salient structure.
In the most basic form, we can consider the problem as a joint optimization of both dimensionality reduction and regression (see Figure~\ref{fig:overview}) that can be formulated as follow:
%Such a design enables a more interpretable pipeline, as a human user can easily find and interpret potentially nonlinear pattern in 2D, but often cannot directly make sense of nonlinear mapping from high-dimensional to 2D. 
%The basic formulation of the optimization can be express as follow: 

%The FPP, as illustrated in Figure~\ref{fig:overview}, employs a join optimization between the linear projection that map high-dimensional space to 2D and the corresponding 2D regressor/classifier.

For a given HD dataset of $N$ samples in $D$ dimensions, $\mathbf{X} \subset \mathbb{X}$, FPP infers $d$-dimensional embeddings $\mathbf{Y} \subset \mathbb{Y}$, based on a response function $f$ defined at each data point $f_i \in \mathbb{F}, \forall i = 1 \cdots N$.
Here, $\mathbb{X}$ and $\mathbb{Y}$ denote the input and the embedded spaces respectively.
The response function space is defined as either $\mathbb{F} \subseteq \mathbb{R}$, in case of continuous response functions, or as $\mathbb{F} \subseteq \Omega$ when $f$ assumes one of $K$ discrete values. This leads to the following general formulation of FPP:
\begin{equation}
\underset{\mathbf{P},  \theta}{\mathrm{argmin}} \frac{1}{N}\sum_{i = 1}^N \mathcal{S}\left[f_i,  g(\mathbf{y}_i ; \theta) \right], \quad \text{where } \mathbf{y}_i = \mathbf{P}^T \mathbf{x}_i.
\label{eq:formulation}
\end{equation}
In this formulation, $g: \mathbb{Y} \mapsto \mathbb{F}$ denotes the mapping function, with parameters $\theta$, between the embedded space and the response function, $\mathbf{P} \in \mathbb{R}^{D \times d}$ is a linear orthonormal projection applied to each data sample $\mathbf{x}_i \in \mathbb{R}^{D}$ and $\mathcal{S}$ is a scoring function used to evaluate the quality of mapping $g$. 
In order to achieve interpretability, FPP relies on linear projections $\mathbf{P}$ and for visualization purposes $d$ is typically fixed at $2$.
One can then capture non-linear relationships by allowing sufficient flexibility for the mapping function $g$, i.e. using higher order regression models. 
For continuous $f$, the examples below use polynomial regressors, where the polynomial degree directly controls the complexity of the inferred structure. 
However, other regressors could easily be integrated as well. 
As scoring function $\mathcal{S}$ any one of the standard goodness of fit measure can be used, such as the mean squared error (MSE). 
In the case of classification when $f$ is discrete, $g$ is defined as a softmax classifier to predict the $K$ class labels. 
%FPP uses simple fully connected layer(s) of a neural network to build the classifier.
%Similarly to the polynomial degree, the number and size of the layers correspond directly to the complexity of the class boundaries. 
In these cases $\mathcal{S}$ is defined as the cross entropy between true and predicted response values. 
Finally, Eq. (\ref{eq:formulation}) can be extended to multiple response functions $f^{l}, l = 1 \cdots L$ as follows:
\begin{equation}
\underset{\mathbf{P},  \{\theta^l\}}{\mathrm{argmin}} \frac{1}{LN} \sum_{l=1}^L\sum_{i = 1}^N \mathcal{S}\left[f^l_i,  g(\mathbf{y}_i ; \theta^l) \right], \quad \text{where } \mathbf{y}_i = \mathbf{P}^T \mathbf{x}_i.
\label{eq:multi}
\end{equation}
Here, our goal is to infer an unified projection $\mathbf{P}$ that simultaneously recovers the $L$ response functions.

To solve Eq.~(\ref{eq:formulation}) requires incorporating the constraint $\mathbf{P}^T \mathbf{P} = \mathbb{I}$ to ensure that the columns of $\mathbf{P}$ are orthonormal and the projection constructs a valid linear subspace. 
More specifically, FPP leverages the popular deep learning framework \emph{TensorFlow}~\cite{tensorflow} to implement a projected gradient descent (PGD) optimization. 
The linear projection is realized with a dense layer with weight matrix of size $\mathbb{R}^{D \times d}$, and the orthonormality constraint is enforced by projecting estimated weights onto the Stiefel manifold, the set of all orthonormal matrices of the form $\mathbb{R}^{D \times d}$, through singular value decomposition (SVD)~\cite{SVD}.
The embeddings from the linear projection, $\mathbf{y} = \mathbf{P}^T\mathbf{x}$, are then used for predicting the response $f$ using a non-linear mapping $g$. 
The detailed steps of FPP is summarized in Algorithm~\ref{algorithm}.
Note that for small $d$, i.e.\ $d=2$ the SVD step is computationally efficient and consequently FPP scales to tens of millions of samples and tens of thousands of dimensions. Our implementation can be found at \url{https://github.com/LLNL/fpp}.

\SetKwInput{KwInput}{Input}                % Set the Input
\SetKwInput{KwOutput}{Output}              % set the Output
\SetKwInput{KwInitialize}{Initialize}  

\begin{algorithm}
	\SetAlgoLined
	\caption{Function Preserving Projections}
	\KwInput{Domain $\mathbf{X} \in \mathbb{R}^{D \times N}$ and response function $f \in \mathbb{R}^{N}$; Scoring function $\mathcal{S}$; Learning rate $\gamma$, mini-batch size $b$}
	
	\KwOutput{Projection matrix: $\hat{\mathbf{P}} \in \mathbb{R}^{2 \times D}$; Parameters $\hat{\theta}$ for $g$}
	
	\KwInitialize{Randomly initialize $\hat{\theta}$ and $\hat{\mathbf{P}}$ (orthonormal matrix)}
	
	\While{exist mini-batch $\widetilde{\mathbf{X}}, \widetilde{f}$ from $\mathbf{X}, \mathbf{f}$ }{
		\vspace{0.5em}
		\tcp{project input HD data onto 2D}
		$\widetilde{\mathbf{y}_i} \xleftarrow{} \hat{\mathbf{P}}^T \widetilde{\mathbf{x}_i}, \forall i = 1 \cdots b$ \;
		\vspace{0.5em}
		\tcp{predict response function and compute goodness of fit}
		$L \xleftarrow{} \sum_{i=1}^b \mathcal{S}[ \widetilde{f}_i, g(\widetilde{\mathbf{y}_i}; \hat{\theta})]$\;
		\vspace{0.5em}
		\tcp{update parameters}
		$\hat{\theta} \xleftarrow{} \hat{\theta} - \gamma \nabla_{\theta}(L)$;
		
		$\hat{\mathbf{P}} \xleftarrow{}\hat{\mathbf{P}} - \gamma \nabla_{\mathbf{P}}(L)$\;
		\vspace{0.5em}
		\tcp{enforce orthonormality constraint}
		
		$\mathbf{U}, \Sigma, \mathbf{V^{T}} \xleftarrow{} \text{SVD}(\hat{\mathbf{P}})$\;  
		$\hat{\mathbf{P}} \xleftarrow{} \mathbf{U}$\;
		
	}
	
	return $\hat{\mathbf{P}}$
		\label{algorithm}
\end{algorithm}

\begin{figure}[!htbp]
\centering
  \includegraphics[width=1.0\linewidth]{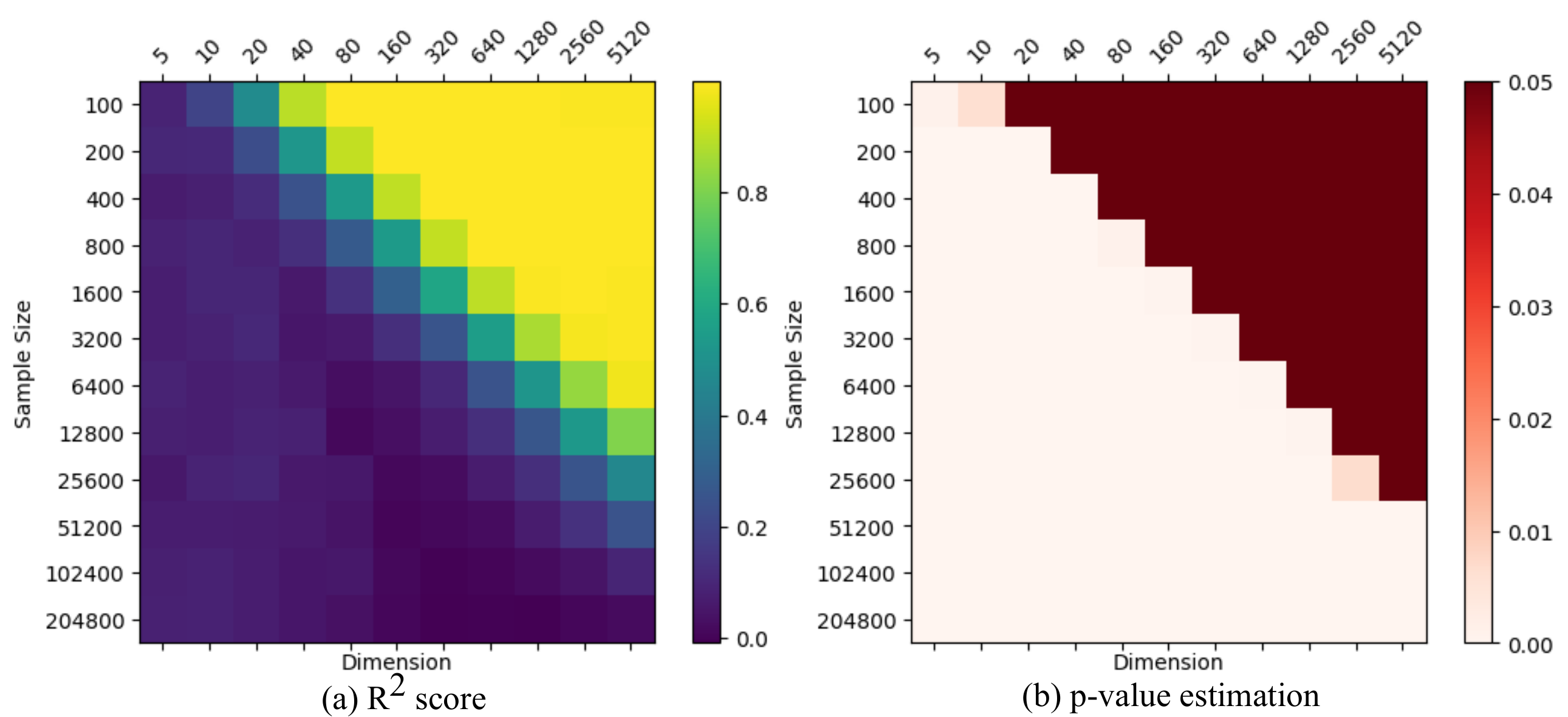}
 \caption{ 
Here we investigate the trustworthiness of the captured pattern in respect to different dimension and sample size combinations.
In this example, we focus on the setup utilizing a 2D polynomial regressor (degree 4).
By examining the potential patterns in totally random data, we can estimate how likely the observed pattern is spurious correlation rather than salient structure. 
In (a), we show the average $R^2$ score of the fitted regressors on random datasets at different conditions. We can then utilize the $R^2$ score samples on random datasets to estimate a p-value for a given data and its corresponding projection. In (b), we show the p-value assuming a projection gives an $R^2$ score of 0.5 for each of the dimension and sample size combinations.
}
\label{fig:pValue}
\end{figure}
%%%%%%%%%%%%%%%%%%%%%%%%%%%%%%%%%%%%%%%%%%%%%%%%%%%%%%%%

For any pattern-finding scheme, it is imperative to evaluate the trustworthiness of the identified pattern. 
Since we are optimizing for low-dimensional patterns in a high-dimensional space, there are many potential opportunities for overfitting. 
We address this challenge from two perspectives:
First, like all statistical problems, we can split the data into training and testing set, fit the projection using training data, and compare the result on both the training and testing set.
If the identified pattern is due to a salient correlation we expect both projections to result in a very similar structure.
Alternatively, we can consider the problem as a hypothesis test, by defining a confidence value (analog to a \emph{p-value}) which describes how likely a pattern of similar strength can be found in random data. 
%For the latter approach, we estimate the value by comparing the score $\mathbf{S}$ from real data with the distribution of the score $\mathbf{S}$ from random data of the same dimension and size. 
Such a test can also provide us with general guidelines on whether we should be concerned about potential overfitting at a given sample count and data dimension. 
As illustrated in Figure~\ref{fig:pValue} (a), for the given 2D polynomial (degree 4) regressor, we show its $R^2$ score when fitted to randomly generated data of different dimension and size.
As expected,  overfitting is more likely to happen as the dimension increases or as sample count decreases. 
Subsequently, we utilize the $R^2$ score samples on random datasets to estimate a p-value for a given projection. 
As illustrated in Figure~\ref{fig:pValue}(b), we show p-values assuming the projection give a $R^2$ score of 0.5 for configurations. 
The colormap is clamped at $p=0.05$, which reveal a clear line separating the significant and non-significant sides. 
Such an observation indicates that a constant factor exists between sample size and dimension size for finding a trustworthy pattern of the regression problem (in this case, the sample size should be at least ten times larger).
We can also obtain the p-value estimation from the loss function value (instead of $R^2$), which can be used for both regression and classification scenarios.
Polynomial regressor is effective for capturing the nonlinear and low-frequency pattern human can easily comprehend. However, the proposed framework does not limit to any particular 2D regressor, provided they are differentiable to utilize the existing SGD framework. 
Also, different types of regressors also impose different priors on the patterns. Therefore, we can utilize the selection of regressor and their setup, i.e., the degree of the polynomial regressor, as a tunable knob in the system to focus on different types of pattern or complexity. However, we should expect to see potentially different overfitting behaviors due to the choice of the model.
For certain datasets with high extrinsic but low intrinsic dimension, one can also utilize random projections as a pre-process to lower the dimensionality of the problem.
By its nature, random projections are highly unlikely to cause overfitting but the resulting lower dimensionality significantly reduces the overfitting risk for the subsequent FPP projection.

%%% Local Variables:
%%% mode: latex
%%% TeX-master: "fpp_arxiv"
%%% End:

\section{Results}
\label{sec:result}
In this section, we demonstrate the applicability of the proposed method on synthetic data as well as on dataset from real-word applications for both regression and classification problems.

\begin{figure}[!htbp]
\includegraphics[width=1.0\linewidth]{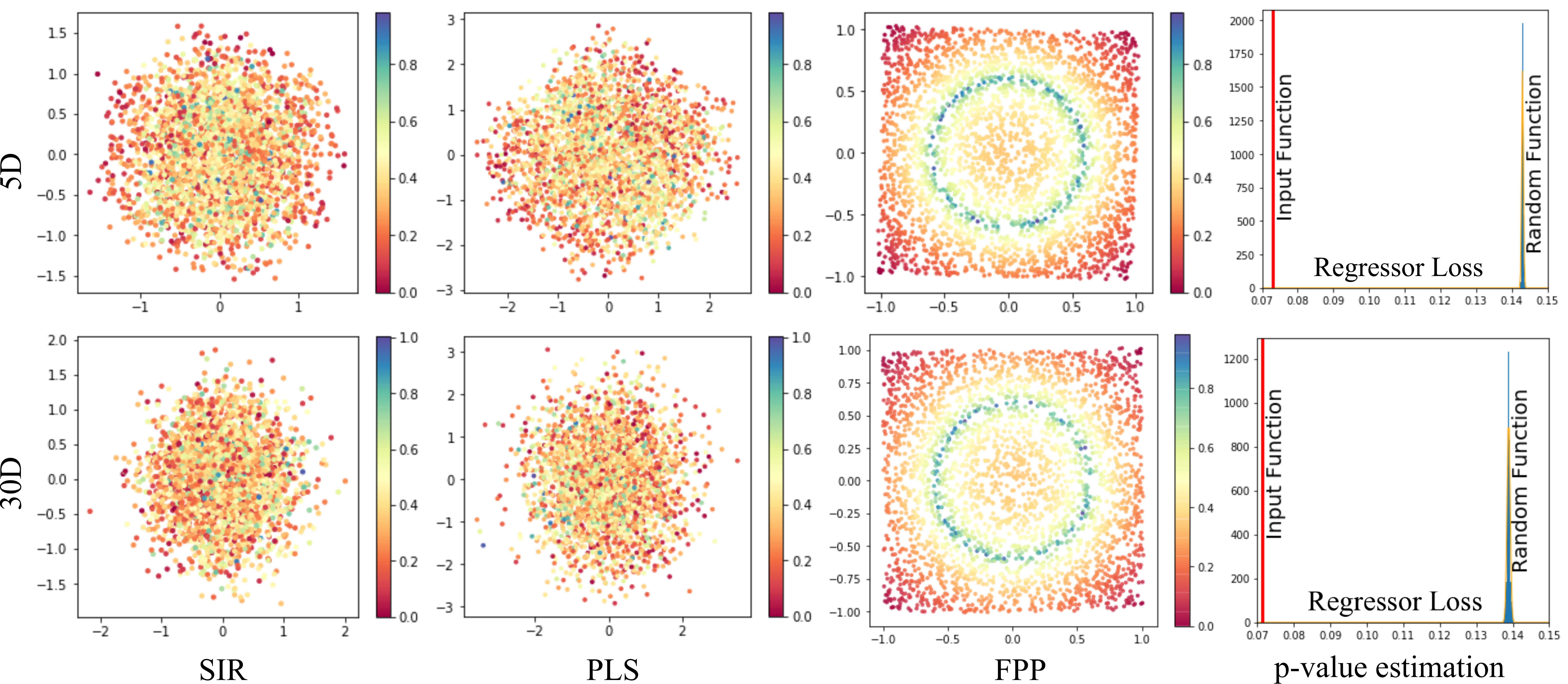}
\caption{
Synthetic data. The top and bottom rows are results from a 5D and 30D function respectively. Here we show that the function preserving projection (FPP) reliably identifies the circle structure in both the 5, 30-dimensional dataset, whereas the partial least square (PLS) and sliced inverse regression (SIR) fail. The rightmost column illustrates the p-value estimation scheme for the observed pattern in the projections. 
Here, we show the distribution of the regression loss values of fitted models on randomized data and then compared it against the loss value obtained from the original function. 
In the plot, the x-axis is the regression loss value. The regression loss for the given function is highlighted by the red line, where as the loss value distribution on the randomized data is shown in blue. The separation between them indicate that the patterns we observed in the projection are statistically significant.
}
\label{fig:synthetic}
\end{figure}

As a synthetic experiment, we define a single function on a uniformly sampled high-dimensional domain. 
Here, the function has a circular pattern (see the third column of the Figure~\ref{fig:synthetic}) in a 2D subspace of a 5D and 30D domain respectively, where each dimension is generated from a uniform random distribution (between $-1.0$ to $1.0$).
In the top row of Figure \ref{fig:synthetic} shows projections from the 5D dataset and the bottom row are from the 30D dataset. 
Due to the linear assumption, both partial least square (PLS) and sliced inverse regression (SIR) fail to capture the circular pattern. By focusing on the visual domain and relying on a nonlinear regressor (in this case, a polynomial regressor of degree 3), the proposed FPP approach can easily reveal the pattern in both the 5D and 30D domain. 
On the rightmost column, we illustrate the significance estimation (p-value) of the pattern captured by the proposed technique. The blue histogram shows the regression loss value distribution for a sample of 300 randomized functions (i.e., a random shuffle of the function values), whereas the red line marker indicates the loss value for the input function. This plot provides insights on whether we overfit to the data by illustrating how likely we will be able to find a pattern of similar strength in the random function. The clear separation between the randomized function regression loss distribution and the input function loss distribution lead to an estimated of p-value 0.0, which indicates a strong evidence against the hypothesis that the observed pattern is from spurious correlation.

\begin{figure}[!htbp]
\centering
  \includegraphics[width=1.0\linewidth]{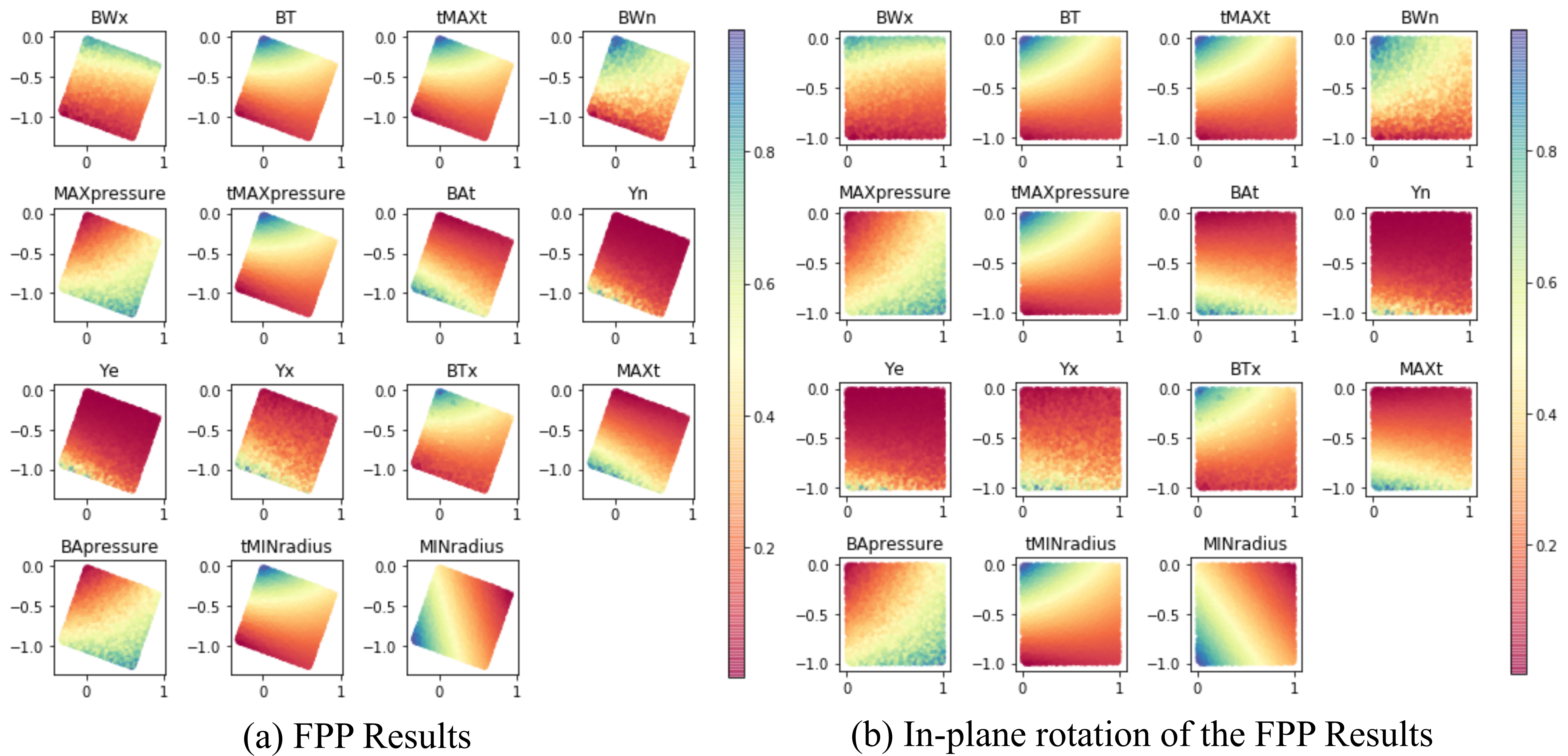}
 \caption{
A demonstration of multi-function projection capability on a physical simulation ensemble dataset.
In this ensemble dataset, we have 5 input parameter and 15 output scalar functions. Here, we aim to identify a single 2D embedding, in which all functions' variation can be explained. The plot on the left shows the subspace identified by FPP, in which all subplots have the same embedding configuration and colored by the values of the 15 outputs, respectively. The configuration can be simplified by making an in-plane rotation, as shown on the right plot. The two directions in these plots correspond to two of the input parameters.
 }
\label{fig:multiFunction}
\end{figure}

In the synthetic dataset, we use FPP to produce a 2D embedding based on one function of interests. 
However, in many applications, we interest in the joint behavior of multiple scalar function, i.e., output properties of physical simulation ensembles.
As discussed in Section~\ref{sec:method}, we can easily extend the single function formulation to multiple ones.
In the following example, we give an example of a multi-function projection, where we produce a single 2D embedding that explains all major variation of the functions.
The application is a physical simulation ensemble (1M samples) produced by a recently proposed semi-analytic simulation model~\cite{gaffney2014thermodynamic, springer2013integrated} for inertial confinement fusion~\footnote{https://github.com/rushilanirudh/icf-jag-cycleGAN}.
The simulator has a 5-dimensional input parameter space and produces several images of the implosion as well as 15 diagnostic scalar outputs. 
For this example, we want to understand what are the main driving factor for not one but all 15 scalar outputs. 
We can explore such a relationship by producing a simple 2D projection that would best capture the changes and pattern of the 15 scalars.
As illustrated in Figure~\ref{fig:multiFunction}(a), by utilizing a degree-3 polynomial regressor for each function, we identify a single 2D projection (color by 15 different scalar values) that explains most of the variation of all scalar outputs in the simulation ensemble.
The fitted model has a loss of 0.0374. To estimate the p-value, we obtain the mean and variance of the loss distribution on randomized function computed from 300 samples, which lead to an estimated p-value of 0.0.
As it returns out, as shown in Figure~\ref{fig:multiFunction}(b), if we apply in-plane rotation, the two dominating directions corresponds to two of the input parameters. According to the physics, the other three parameters (out of five input parameters) are shape parameters, therefore, they will mostly impact the generated image instead of the scalars. This example provides a real word verification of the proposed ability to capture a shared configuration, which helps interpret a set of functions defined in the same domain.

%%%%%%%%%%%%%%%%%%%% classification %%%%%%%%%%%%%%%%%%%%%%%%%

\begin{figure}[!htbp]
\centering
  \includegraphics[width=1.0\linewidth]{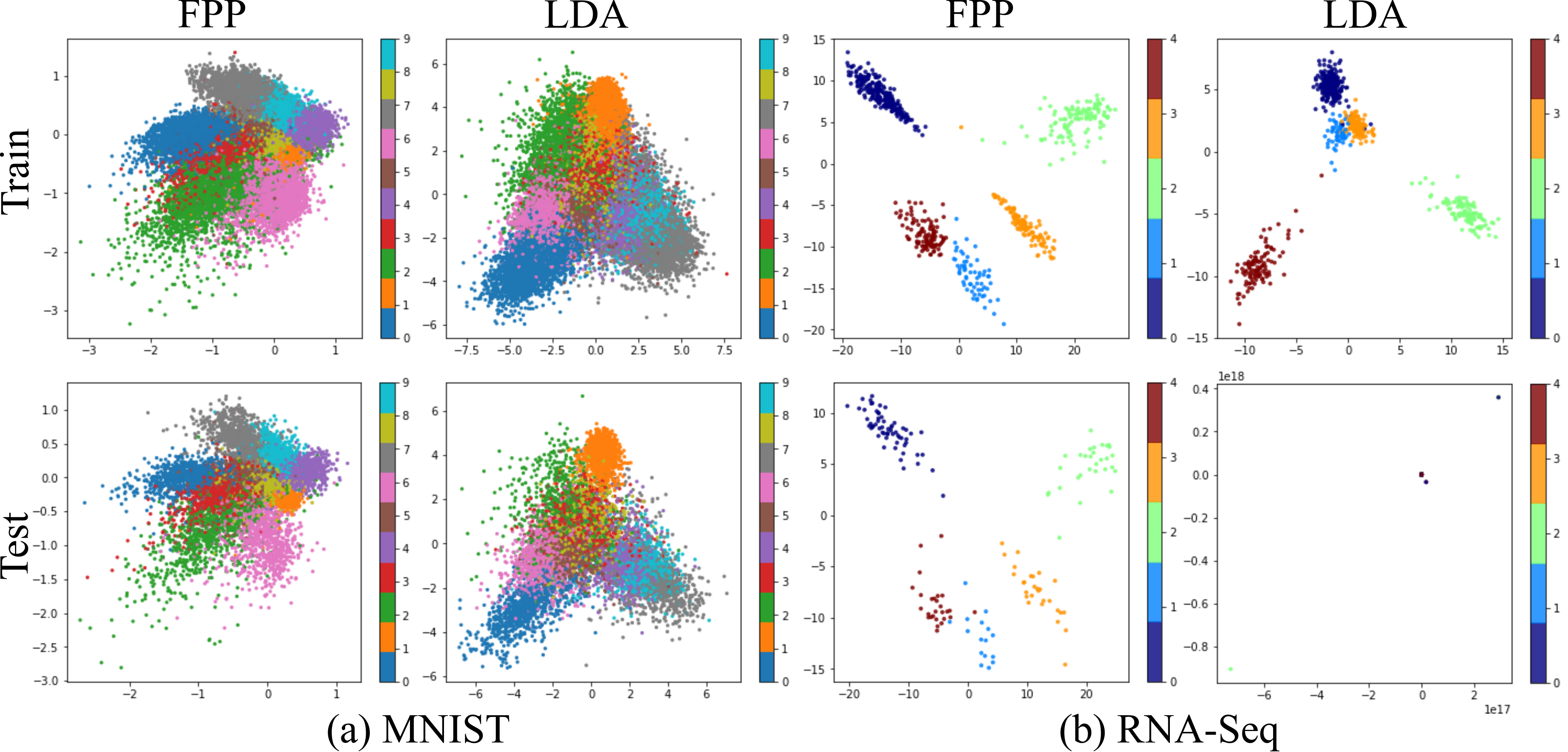}
 \caption{
 FPP projection examples where the function is the class labels. In (a), the MNIST image data with 10 labels is illustrated, whereas the RNA sequence (RNA-seq) data with 5 labels is shown in (b). The proposed method generates more clearly separation for each class for both data. In particular, the proposed method also produces a more robust projection for the RNA-seq example, where the LDA is overfitted to the training data and fails to generalize to test set.
 }
\label{fig:classificationExample}
\end{figure}

In previous examples, we have demonstrated the effectiveness of FPP for projecting continuous, high-dimensional functions, i.e., for regression problems.
For classification problems, i.e., find the 2D projection that best separate samples with different labels, we can simply replace the 2D regressor with a 2D classifier. 
Here, we utilize a 2D logistic regression classifier (with an additional nonlinear layer) to drive the selection of the linear projection. 
In Figure \ref{fig:classificationExample}(a), we compare the proposed FPP with linear discriminant analysis (LDA)~\cite{LDA} on the MNIST dataset~\footnote{http://yann.lecun.com/exdb/mnist/} that consists of 60K sample of 28 by 28 images. 
We can see the FPP can find a 2D linear projection that separates different digits' images better than the LDA result. 
We also apply the method on a high-dimensional RNA sequence data~\footnote{https://archive.ics.uci.edu/ml/datasets/gene+expression+cancer+RNA-Seq}, which has more than 20531 feature dimensions and 801 samples. 
Due to the extremely high-dimensional and the very low sample count, there is a high potential for overfitting. 
As discussed in previous examples, we can obtain the p-value $2.027\mathrm{e}{-07}$, which give high confidence on the captured structure.
%As demonstrated in the synthetic data example, we can utilize p-value  (see details in Section~\ref{sec:method}) to estimate how likely we will fit a model with a similar error when the data is random. 
%Such an approach represents the minimal criteria for validation, which is better at providing a general guideline than presenting a definitive answer on model overfitting. 
%\ptb{This sounds like an excuse .. we should at least provide the p-value or take this sentence out.}
Alternatively, we can validate the trustworthiness of both the FPP and LDA results by split the datasets into training and test and then evaluate the trained models on the test set.
As shown in Figure \ref{fig:classificationExample}(b), not only does FPP separate the class better than LDA, but it also produces more robust projection that generalized well to the test set, unlike the LDA projection which entirely fails on the test set.
When examining the LDA projection matrix, we notice only 3 of the 20K dimensions are contributing to the projection, which leads to a degenerate projection for the test set.

%%%% ImageNet %%%%

\begin{figure}[!htbp]
\centering
  \includegraphics[width=1.0\linewidth]{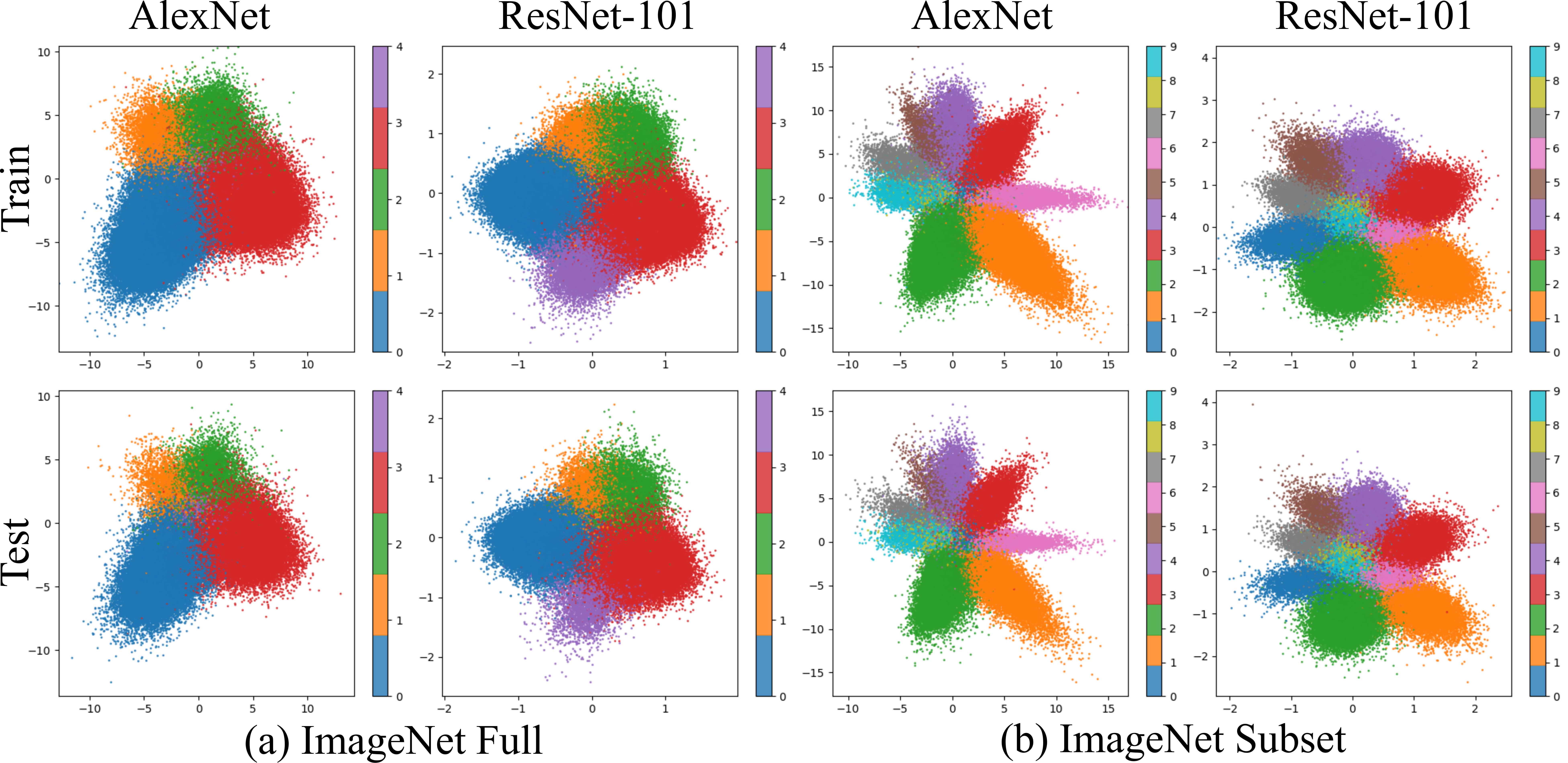}
 \caption{
Projections of imageNet image feature representations from different neural network architectures. From the comparison, we can see that the representation produced by ResNet-101 is easier to separate by categories in the linear projection than that of the AlexNet.
}
\label{fig:imageNet}
\end{figure}

The 2D loss function combined with the SGD based implementation allows FPP to scale to data sizes significantly beyond the ability of most traditional projection/dimensionality reduction approaches. 
In the following example, we use FPP to probe into feature representations of two popular deep learning architecture (AlexNet, ResNet) on the entire ImageNet challenge~\cite{imageNet} training dataset, which consists of more than 1.28 million images and 1000 classes.
We use the last layer before the \emph{softmax} as the feature representation for both networks resulting in a 2048- and 4096-dimensional feature space for ResNet and AlexNet respectively.  Combined with a large number of samples this results in datasets of more than 40GB.
Despite their massive size, we can generate projections for each of these 1.28M sample datasets within minutes (between 3 to 10 minutes). The detailed timing results and parameter setups for all our examples are in listed in Table~\ref{tab:performance}. For the first five datasets, we compute the results on a laptop with an Intel Core i7-6820HQ processor (2.9GHz), whereas the last four are computed on a server (due to limited memory on the laptop) with an Intel Xeon E5-2695 processor (2.1GHz).

For the experiments, we first group the entire 1.28M images (1000 classes) into 5 coarse categories, namely, ``living thing'', ``natural object'', ``food'', ``artifact'', ``misc'', where the first 4 categories consists 984 of the 1000 classes available in the imageNet challenge. We then project all the images feature representations with the categories as the label.
As shown in Figure~\ref{fig:imageNet}(a), we can see that the AlexNet's representation has trouble distinguishing the purple samples (``misc'' category) from the rest, whereas the ResNet's representation, despite being lower-dimensional, can.
For visualizing more detailed category separability, we generate 10 categories with more concrete and meaningful labels (i.e., ``fish'',  ``bird'', ``mammal'', ``invertebrate'', ``food'', ``fruit'', ``vehicle'', ``appliance'', ``tool'', ``instrument''), which consists of 611k images of the 1.28M.
As we can see in Figure~\ref{fig:imageNet}(b), the feature representation of ResNet again seems to be able to better separate these categories compared to AlexNet's representations, which may explain the gap in their predictive performances (AlexNet and ResNet have Top-5 errors of 20.91\% and 6.44\%, respectively).

%%%%%%%%%% performance discussion %%%%%%%%%
\begin{table}
\caption{Performance}
\label{tab:performance}
\begin{tabular}{ p{3.3cm}||p{1.0cm}|p{1.0cm}|p{0.8cm}|p{1.6cm} |p{0.8cm} |p{0.8cm} |p{1.1cm} }
 \hline
 Dataset & sample & domain & range & size & epoch & batch & timing(s) \\
 \hline
Circle (5D)   & 3000  & 5 & 1 &  0.14 (MB) & 50 & 50 & 2.98 \\
Circle (30D) &  3000  & 30  & 1 & 0.74 (MB) & 50 & 50 & 3.39 \\
ICF simulator    & 1M & 5 & 15 & 152.6 (MB) & 1 & 200 & 24.7 \\
RNA-Seq &   801  & 20531 & 1 &  125.5 (MB) & 10 & 30 & 1.18 \\
MNIST    & 60K & 784  & 1 &  179.4 (MB) &  10 & 100 & 9.57 \\
ResNet ImageNet Sub & 611K  & 2048 & 1 & 9.3 (GB) & 10 & 200 & 184.5 \\
ResNet ImageNet Full & 1.28M  & 2048 & 1 & 19.6 (GB) & 10 & 200 & 404.8 \\
AlexNet ImageNet Sub & 611K  & 4096 & 1  & 18.7 (GB) & 10 & 200 & 283.7 \\
AlexNet ImageNet Full & 1.28M  & 4096 & 1  & 39.1 (GB) & 10 & 200 & 640.1\\
 \hline
\end{tabular}
\end{table}
%%% Local Variables:
%%% mode: latex
%%% TeX-master: "fpp_arxiv"
%%% End:

\section{Conclusion}
In this work, we introduce a novel class of linear projection methods for visualizing interesting and interpretable visual patterns of the function in 2D subspaces of the function domain. The combination of linear projection and nonlinear pattern searching schemes (i.e., a polynomial regressor, or a nonlinear classifier in 2D) allows us to exploit our innate ability to perceive complex (and potentially nonlinear) visual pattern in 2D while compensating our inability to comprehend nonlinear transformation by focus only on a linear transformation from high-dimensional space to 2D. The efficient formulation also allows us to easily scale the problem beyond million of samples and tens of thousands of dimensions that is unimaginable for most dimensionality reduction methods.

 \subsubsection*{Acknowledgments}
This work was performed under the auspices of the U.S. Department of Energy by Lawrence Livermore National Laboratory under Contract DE-AC52-07NA27344. Released under LLNL-JRNL-790959.

% Use unnumbered third level headings for the acknowledgments. All acknowledgments
% go at the end of the paper. Do not include acknowledgments in the anonymized
% submission, only in the final paper.

\bibliographystyle{plainnat}
\bibliography{HDFproj}

\newpage
\appendix
%\section{Appendix}
\section{Hypothesis Testing Failure Case}

\begin{figure}[!htbp]
\centering
  \includegraphics[width=1.0\linewidth]{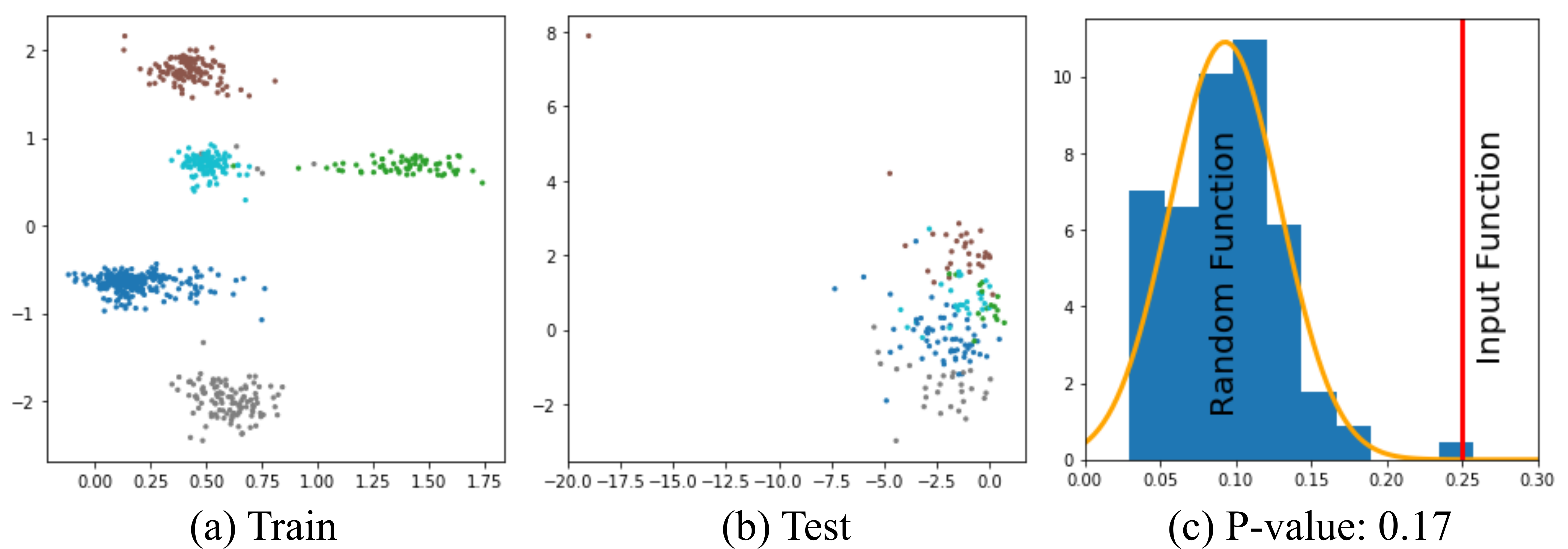}
 \caption{
Overfit example. As shown in (a), the FPP finds a 2D projection that separate classes in the training data. However, we can confirm the existence of overfitting behavior by estimating the p-value (0.17) as shown in (c). We can also observe the failure by projecting the test data as illustrated in (b).
}
\label{fig:overfit}
\end{figure}

As illustrated in Figure~\ref{fig:pValue}, when the ratio between data sample and dimension count is relatively large (i.e., more than 20), we often do not need to worry about overfitting the model to spurious correlation in the data. This observation is also supported by the examples in the result, where we always obtain very small p-values when the sample to dimension ratio is high. 
In the following example, we illustrate a scenario, where the model overfits to the data. We show that both the p-value and the train/test split help reveal the problem.
Here we again look at the RNA-seq dataset (20531 dimensions, 801 samples). Based on the sample and dimension ratio, there is a high likelihood of overfitting for a regression problem. 
In this example, instead of using a 2D classifier (as in the original result), we treat the labels as multiple binary functions and solve it using the multi-function regression setup.
As illustrated in Figure~\ref{fig:overfit}(a), we are able to separate the 5 classes even in the regression setup. However, when estimating the p-value from that projection, as shown in (c), we obtain an estimated value of 0.17, which indicates an untrustworthy result. The overfitting problem can also be detected by projecting the test dataset (b).

%%%%%%%%%%%%%%%%%%%%%%%%%%%%%%%%%%%%%%%%%%%%%%%%%%%%%%%%
\section{Additional Details For the ImageNet Results}

%\begin{figure}[!htbp]
%\centering
%  \includegraphics[width=1.0\linewidth]{imageNet}
% \caption{
%Projections of imageNet image feature representations from different neural network architectures. From the comparison, we can see that the representation produced by ResNet-101 is easier to separate by categories in the linear projection than that of the AlexNet.
%}
%\label{fig:imageNet}
%\end{figure}

\begin{figure}[!htbp]
\centering
  \includegraphics[width=1.0\linewidth]{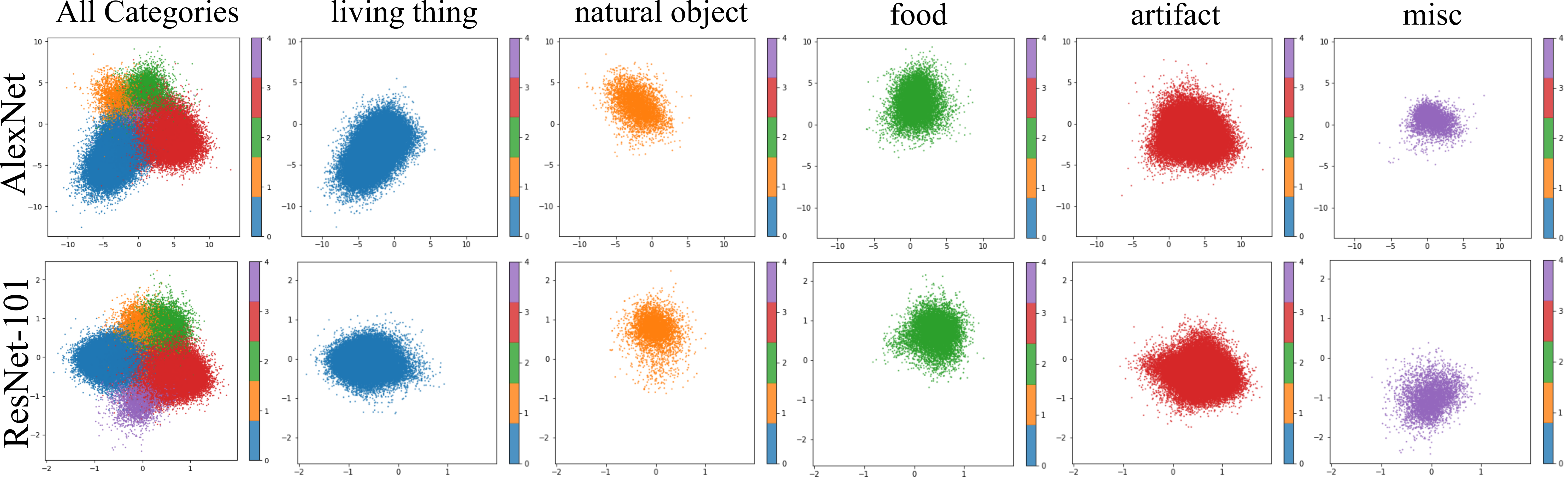}
 \caption{
Decomposition of the multi-category plots of the full imageNet dataset projections with 5 categories.
}
\label{fig:decomposition}
\end{figure}

\begin{figure}[!htbp]
\centering
  \includegraphics[width=1.0\linewidth]{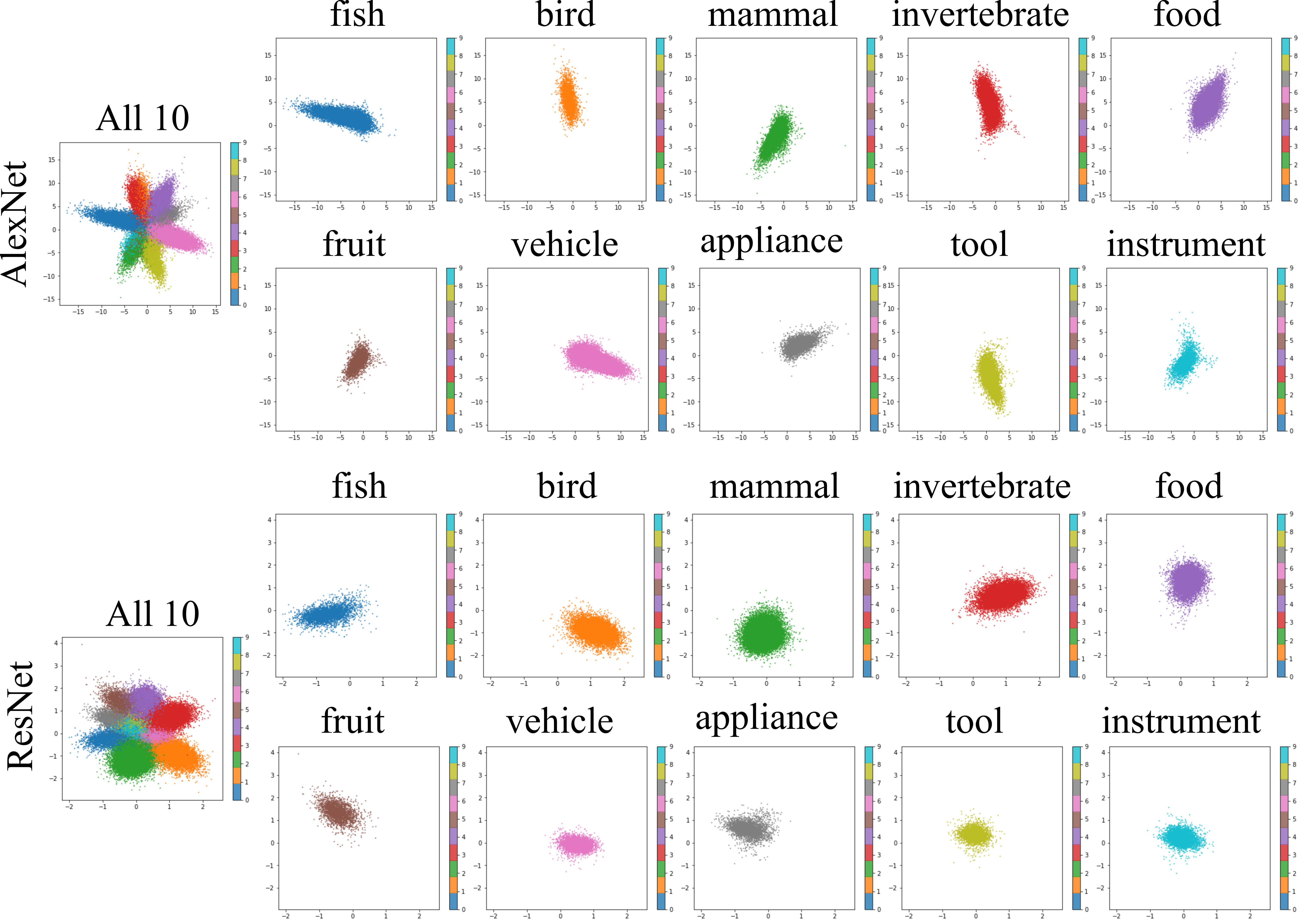}
 \caption{
Decomposition of the multi-category plots of the subset imageNet dataset projections with 10 categories.
}
\label{fig:decomposition2}
\end{figure}

In the result section, we illustrate category separation of the full and subset of the imageNet dataset. However, due to the occlusion caused by a large number of samples in the 2D plots, we may not have a clear understanding of the distribution for each of the category.
Here in Figure~\ref{fig:decomposition} and Figure~\ref{fig:decomposition2}, we show the class distribution details by decomposing each multi-category plot into multiple plots where only samples from one category are displayed.

% \section*{References}
%
% References follow the acknowledgments. Use unnumbered first-level heading for
% the references. Any choice of citation style is acceptable as long as you are
% consistent. It is permissible to reduce the font size to \verb+small+ (9 point)
% when listing the references. {\bf Remember that you can use more than eight
%   pages as long as the additional pages contain \emph{only} cited references.}
% \medskip
%
% \small
%
% [1] Alexander, J.A.\ \& Mozer, M.C.\ (1995) Template-based algorithms for
% connectionist rule extraction. In G.\ Tesauro, D.S.\ Touretzky and T.K.\ Leen
% (eds.), {\it Advances in Neural Information Processing Systems 7},
% pp.\ 609--616. Cambridge, MA: MIT Press.
%
% [2] Bower, J.M.\ \& Beeman, D.\ (1995) {\it The Book of GENESIS: Exploring
%   Realistic Neural Models with the GEneral NEural SImulation System.}  New York:
% TELOS/Springer--Verlag.
%
% [3] Hasselmo, M.E., Schnell, E.\ \& Barkai, E.\ (1995) Dynamics of learning and
% recall at excitatory recurrent synapses and cholinergic modulation in rat
% hippocampal region CA3. {\it Journal of Neuroscience} {\bf 15}(7):5249-5262.

\end{document}